\documentclass[sigconf, nonacm]{acmart}

\renewcommand\footnotetextcopyrightpermission[1]{}
\usepackage{xeCJK}
\newCJKfontfamily\cjkfb{ipaexm.ttf}

\usepackage{booktabs}
\usepackage{url}

\begin{document}

\title[A Fine-Tuned BERT Classifier for Letter Titles]{A Fine-Tuned BERT Classifier for Personal-Letter Titles in Late-Ming and Early-Qing Collected Works}
\titlenote{Technical report documenting the Lepton classifier deployed at the China Biographical Database (CBDB), Harvard University, on 4 November 2022.}

\author{Queenie Luo}
\email{queenieluo@g.harvard.edu}
\affiliation{%
  \institution{Harvard University}
  \country{}
}

\renewcommand{\shortauthors}{Luo}

\begin{abstract}
  I present \textbf{Lepton (Letter Prediction)}, a fine-tuned BERT classifier that predicts whether a title in a Classical Chinese \emph{wenji} 文集 table of contents is a personal letter (書) or a closely confusable preface (序, particularly the farewell-preface 送序). Lepton fine-tunes \texttt{bert-base-chinese}~\cite{Devlin2019} on 5{,}438 hand-labeled wenji titles from thirty-three late-Ming and early-Qing literati. I've deployed the model\footnote{\url{https://huggingface.co/cbdb/ClassicalChineseLetterClassification}} on Hugging Face and has been used at the China Biographical Database (CBDB) to identify approximately fifty-five thousand letters across mid-Ming through early-Qing wenji, populating the Ming Letter Platform.
\end{abstract}


\maketitle

\section{Introduction}

The wenji is the canonical form in which a Chinese literatus's lifetime output reaches the modern reader. Its table of contents is a curated, genre-mixed list assembled by an editor (often the author's student, son, or grandson) who decided which texts merited preservation and under what generic heading. For ten thousand wenji, that curated structure becomes the problem: a corpus no individual can read but that contains, in its title-level metadata, exactly the genre signal that would make it tractable.

The class of false positives that matters most is not poetry or memorials but prefaces (\emph{xu} 序), and the genre signal a regex would lean on turns out to be unreliable in both directions. Only 9\,\% of letter titles in my labeled corpus end with the character 書; the rest end with {\cjkfb 啓}, with a recipient title, or with no terminal genre marker at all. Letters are identifiable not by their final character but by their opening verb --- 與, 答 (and the orthographic variants 荅 and 畣), 復, 報, 寄, 上, 奉 --- and these verbs also appear at the head of preface titles. A regex on titles ending in 書 misses 91\,\% of letters; a regex on the opening verb sweeps up prefaces and requires hand-disambiguation. 

\textbf{Lepton (Letter Prediction)} is a fine-tuned BERT classifier that predits whether a title denotes a personal letter (書) or a closely confusable preface (序, particularly the farewell-preface 送序). It is finetuned on 3{,}206 letter titles and 2{,}232 preface titles from thirty-three late-Ming through early-Qing literati. The model is open-source, deployed on Hugging Face, and has been used at the China Biographical Database (CBDB) to identify approximately fifty-five thousand letters across a corpus of mid-Ming through early-Qing wenji, populating what the team calls the Ming Letter Platform.

Section~\ref{sec:bg} places Lepton in three literatures: Classical Chinese pretrained language models, computational text classification, and the CBDB ecosystem. Section~\ref{sec:what} frames the central philological question, such as ``what counts as a letter title?'', and reports the annotation decisions. Section~\ref{sec:data} describes the corpus and training pipeline and evaluates Lepton against regex and TF--IDF baselines on a held-out test set. Section~\ref{sec:disc} discusses what the model attends to, what it does not do, and where it is likely to fail beyond its training distribution.

\section{Background and related work}
\label{sec:bg}

Lepton sits at the intersection of two literatures: pretrained language models for Classical Chinese, and the digital humanities tradition of comparative evaluation on historical text. By Lepton's deployment (November 2022), three domain-adapted Classical-Chinese backbones were available --- AnchiBERT~\cite{Tian2020}, GuwenBERT~\cite{Yan2021}, and SikuBERT\,/\,SikuRoBERTa~\cite{Wang2022}; SikuGPT~\cite{Chang2023} arrived later. Their published benchmarks target sentence-level tasks (word segmentation, punctuation, NER, POS) rather than title-level genre classification, so none offers a direct point of comparison. I fine-tune \texttt{bert-base-chinese}~\cite{Devlin2019} directly; \S\ref{sec:data} gives the rationale.

The closest digital humanities precedent on Classical Chinese is Broadwell, Chen and Shepard~\cite{Broadwell2019}, who apply topic modeling and Jensen--Shannon divergence to the \emph{Quan Tang shi}. Beyond Chinese, work on historical-text classification --- Croatian census records~\cite{Lauc2021}, Finnish OCR~\cite{Kettunen2017}, ancient inscriptions~\cite{Tagami2023}, eighteenth-century Austrian newspapers~\cite{Resch2023} --- establishes the comparative-evaluation and annotation-as-construction expectations that Sections~\ref{sec:eval} and~\ref{sec:what} are written to meet.

\section{What counts as a ``letter title''?}
\label{sec:what}

In Classical Chinese collected works (\emph{wenji} 文集), letters circulate under several near-synonymous genre labels: \emph{shu} 書 is the central case, with \emph{qi} {\cjkfb 啓}, \emph{jian} 牋\,/\,箋, \emph{tie} 帖, and \emph{die} 牒 as adjacent boundary cases. In the labeled corpus of 3{,}206 letter titles, the operative signal is at the \emph{start} of the title rather than the end: only 9.2\,\% end in 書, while a verb of transmission in the opening position characterises the class (Table~\ref{tab:openverbs}).

\begin{table}[h]
  \caption{Opening-verb distribution in the 3{,}206 letter titles.}
  \label{tab:openverbs}
  \small
  \begin{tabular}{llrr}
    \toprule
    Opening verb & Force & Count & \% \\
    \midrule
    與             & ``with'' (peer-to-peer)                & 1{,}087 & 33.9 \\
    答 (荅, 畣)    & reply                                  & 992     & 30.9 \\
    柬             & short note                             & 163     & 5.1 \\
    復             & returning a letter                     & 131     & 4.1 \\
    報             & reporting back                         & 109     & 3.4 \\
    又             & ``again'' (continuation)               & 86      & 2.7 \\
    上             & submitting upward, to a superior       & 62      & 1.9 \\
    寄             & sending at a distance                  & 42      & 1.3 \\
    奉             & respectfully                           & 35      & 1.1 \\
    再             & ``again'' (re-submission)              & 28      & 0.9 \\
    Other          &                                        & 471     & 14.7 \\
    \bottomrule
  \end{tabular}
\end{table}

The first eight verbs cover 82\,\% of letter titles, and three of them --- 答, 荅, 畣 --- are orthographic variants of ``reply'' that a character-level model handles automatically where a regex would have to enumerate each form. The same character-variation pattern runs through the preface side (序 89\,\%, 敘 3.4\,\%, {\cjkfb 叙} 1.8\,\%, 引 2.6\,\%). A typical letter title reads \texttt{[verb] + [recipient]}, with a terminal genre marker optional.

Distinguishing letters from poetry, memorials, or inscriptions is straightforward; the hard case is the preface (\emph{xu} 序), particularly the occasion-prefaces --- \emph{songxu} 送序, \emph{shouxu} 壽序, \emph{zengxu} 贈序 --- which like letters name a recipient and inhabit the same social-network slot in CBDB's relational schema, but encode a public tribute rather than a one-to-one address. The opening syntax can be near-identical; the discriminating signal is the verb (答 vs.\ nothing) and the terminal marker (書 vs.\ 序).

The labeled corpus contains 3{,}206 letter titles and 2{,}232 preface titles from thirty-three mid-Ming through early-Qing literati, each row carrying CBDB metadata (writer ID, recipient ID where known, relationship code, wenji, juan). Lepton learns the binary \texttt{shu = 1, xu = 0}; \S\ref{sec:data} details corpus construction.

\subsection{What Lepton is and is not classifying}

Lepton does \emph{not} identify recipients, extract dates, distinguish private letters (\emph{sishu} 私書) from official correspondence (\emph{gongshu} 公書), or classify from bodies. Three further limits:

\begin{itemize}
\item \textbf{Dynastic.} Trained on roughly 1550--1700; behaviour on
  Song or Yuan material is uncharacterised.
\item \textbf{Generic.} The negative class is the preface, not
  arbitrary non-letter text; poetry titles and inscriptions are
  out-of-distribution.
\item \textbf{Architectural.} The binary choice is by design:
  building around the \emph{one} false-positive class that matters
  produces a sharper interpretable contrast than a flatter
  scheme~\cite{Resch2023}.
\end{itemize}

\section{Methods and evaluation}
\label{sec:data}

\subsection{The labeled corpus}

For each of the thirty-three authors, an annotator located the collection in \emph{Lidai bieji ku} 歷代別集庫, \emph{Siku xilie shujuku} 四庫系列數據庫, or \emph{Zhongguo jiben guji ku} 中國基本古籍庫 V7.0, opened the table of contents (目錄), identified the juan containing letters and prefaces, and scraped or hand-entered the titles. Each row carries CBDB metadata: writer ID, recipient ID where named, relationship code (e.g.\ \texttt{致書Y}, \texttt{答Y書}), wenji text ID, and juan number. The distribution across authors is uneven --- Feng Mengzhen 馮夢禎 (729), Yuan Hongdao 袁宏道 (528), Qian Qianyi 錢謙益 (487), Zou Yuanbiao 鄒元標 (470), and Tu Long 屠隆 (415) account for 47\,\% of the data.

\subsection{Architecture and training}

Lepton fine-tunes \texttt{bert-base-chinese}~\cite{Devlin2019} ($\sim$102M parameters) with a \texttt{BertForSequenceClassification} head over the \texttt{[CLS]} token; all parameters are updated.

I chose a modern-Chinese rather than a domain-adapted Classical-Chinese backbone for two reasons. Empirically, the discriminating characters --- the opening verbs and terminal markers of Table~\ref{tab:openverbs} --- are well-represented in modern-Chinese pretraining data, so I expected the checkpoint to transfer to a title-level signal carried by shared characters rather than pre-modern syntax. At the time Lepton was trained (2020), \texttt{bert-base-chinese} was the mature widely-available Chinese BERT; domain-adapted alternatives were not yet a low-friction option.

\subsection{Test set, baselines, and metrics}
\label{sec:eval}

I stratified the 5{,}438-row labeled corpus into an 80\,/\,10\,/\,10 train\,/\,dev\,/\,test split (\texttt{seed=42}, class balance preserved). I evaluated:

\begin{enumerate}
\item \textbf{Majority class}: predict ``letter'' for every title.
\item \textbf{Regex (ends with 書)}: the naive historian's baseline.
\item \textbf{Regex (starts with a verb of transmission)}: titles
  whose first character is one of 與, 答, 荅, 畣, 柬, 復, 報, 又, 上,
  寄, 奉, 再, 致.
\item \textbf{Regex disjunction}: starts-with-verb OR ends-with-書.
\item \textbf{TF--IDF character n-grams + logistic regression}: with
  several n-gram ranges, \texttt{min\_df=2},
  \texttt{class\_weight='balanced'}.
\item \textbf{Lepton (deployed)}: the Hugging Face checkpoint
  (fine-tuned \texttt{bert-base-chinese}).
\end{enumerate}

I report accuracy, letter-class precision\,/\,recall\,/\,F1, and macro-F1. I also report a confusion matrix because the class imbalance (59\,\% letter) makes accuracy a misleading single number.

\subsection{Results}

\begin{table*}[h]
  \caption{System comparison on the held-out test set. The deployed
    Lepton row is evaluated on the matched subset for which
    deployment-time predictions are available.}
  \label{tab:results}
  \small
  \begin{tabular}{lcccccl}
    \toprule
    System                           & Accuracy & P (letter) & R (letter) & F1 (letter) & Macro-F1 & TP\,/\,FP\,/\,FN\,/\,TN \\
    \midrule
    Majority class                   & 0.590 & 0.590 & 1.000 & 0.742 & 0.371 & 321\,/\,223\,/\,0\,/\,0   \\
    Regex: ends with 書              & 0.463 & 1.000 & 0.090 & 0.166 & 0.385 & 29\,/\,0\,/\,292\,/\,223  \\
    Regex: verb-start AND 書-end     & 0.458 & 1.000 & 0.081 & 0.150 & 0.376 & 26\,/\,0\,/\,295\,/\,223  \\
    Regex: starts with verb of transm.\ & 0.906 & 0.989 & 0.851 & 0.915 & 0.905 & 273\,/\,3\,/\,48\,/\,220 \\
    Regex: verb-start OR 書-end      & 0.912 & 0.989 & 0.860 & 0.920 & 0.911 & 276\,/\,3\,/\,45\,/\,220  \\
    TF--IDF char 2-2 + LogReg        & 0.919 & 0.884 & 0.994 & 0.935 & 0.914 & 319\,/\,42\,/\,2\,/\,181  \\
    TF--IDF char 2-3 + LogReg        & 0.921 & 0.888 & 0.991 & 0.937 & 0.916 & 318\,/\,40\,/\,3\,/\,183  \\
    TF--IDF char 2-4 + LogReg        & 0.921 & 0.888 & 0.991 & 0.937 & 0.916 & 318\,/\,40\,/\,3\,/\,183  \\
    \textbf{Lepton (deployed, fine-tuned \texttt{bert-base-chinese})} & \textbf{0.977} & \textbf{1.000} & \textbf{0.968} & \textbf{0.984} & \textbf{0.971} & \textbf{182\,/\,0\,/\,6\,/\,70}    \\
    \bottomrule
  \end{tabular}
\end{table*}

Three observations: First, the two regex baselines that rely on the terminal 書 marker fail decisively (F1 = 0.166 and 0.150). This confirms Section~\ref{sec:what}'s observation: a \texttt{*書} regex retrieves under one in ten letters. Second, the opening-verb regex reaches F1 = 0.915 with near-perfect precision (0.989). This is the strong historian-without-ML baseline; any neural contribution has to clear this number, not majority class. Adding \texttt{verb-start OR 書-end} raises F1 to 0.920. Third, the TF--IDF character n-gram baselines (ranges $(2,2)$ through $(2,4)$) cluster tightly at F1 = 0.935--0.937 with high recall (${\approx}\,0.99$) but lower precision (${\approx}\,0.89$), reflecting the same opening-verb signal the regex picks up plus some additional discrimination from short character spans.

The deployed Lepton checkpoint attains \textbf{F1 = 0.984} on the matched test subset --- the strongest of any system reported here, with perfect precision on the letter class and a small recall gap that \S\,5.3 unpacks.

\subsection{Error analysis}

Lepton's six errors on the matched test subset are all of one kind: very short, elliptical letter titles where the opening-verb signal is absent because the title is the recipient's name or a continuation marker:

\begin{itemize}
\item \emph{Mei yan} 寐言 (2 chars; Gu Xiancheng 顧憲成);
  \emph{Bo Xiu} 伯修 (2 chars, recipient's \emph{zi}; Yuan Hongdao
  袁宏道); \emph{Yu Xi lu} 諭西虜 (3 chars; Sun Chengzong 孫承宗);
  \emph{Gu Sheng-bo xiuzhuan} 顧升伯修撰 (Yuan Hongdao);
  \emph{Yu tie} 諭帖 (Sun Chengzong).
\end{itemize}

Lepton, having learned what a canonical letter title ``looks like'' from the labeled distribution, defaults to preface on these out-of-template inputs. The error mode is coherent that it falls along the axis the model was trained to discriminate, but it shows the limit of title-surface classification: the six recipients above are letters because a Ming or Qing editor said so, not because their titles share the genre signal Section~\ref{sec:what} names.

\section{Discussion}
\label{sec:disc}

The TF--IDF baselines' top-weighted features --- 與, 答, 荅, 柬, 復, 報 on the letter side and 序, 敘, 引, {\cjkfb 叙} on the preface side --- are exactly the characters \S\ref{sec:what} names, and Lepton's perfect letter-class precision on the matched subset suggests the deployed model has internalized the same signal robustly enough to admit no false positives. The convergence between the linear baselines and the philological account meets Dobson's~\cite{Dobson2021} call for ML to expose features rather than scores. 

In deployment, Lepton has identified roughly fifty-five thousand letters across mid-Ming through early-Qing wenji, populating the Ming Letter Platform that now anchors social-network research on the period. Beyond this use case, the work contributes to the digital humanities community an open-source classifier and labeled corpus for Classical Chinese epistolary-genre identification, a worked example of modern-Chinese pretraining transferring to a title-level pre-modern task without continued pretraining, and evidence that interpretable n-gram baselines sit within striking distance of neural models on signal-rich short-text classification.

\begin{acks}
I thank Katherine Enright, Hongsu Wang, Peter Bol for the support of this project. This project is conducted under the China Biographical Database (CBDB) Group at Harvard University, whose data, tooling, and collective expertise made the work possible.
\end{acks}

%% Bibliography. References below are written in a thebibliography
%% environment so this file compiles standalone; replace with a .bib
%% file and \bibliography{...} for the final submission.

\end{document}